\documentclass[letterpaper, 10 pt, conference]{template/ieeeconf}  

\IEEEoverridecommandlockouts                              

\usepackage{graphics} 
\usepackage{epsfig} 
\usepackage{mathptmx} 
\usepackage{times} 
\usepackage{amsmath} 
\usepackage{amssymb}  
\usepackage{booktabs}
\usepackage{makecell}
\usepackage{multirow}
\usepackage{siunitx}
\usepackage{authblk}

\title{\LARGE \bf
Fine-Tuning Foundation Models with Federated Learning for Privacy Preserving Medical Time Series Forecasting
}

\author[1]{Mahad Ali}
\author[2]{Curtis Lisle}
\author[3]{Patrick W. Moore}
\author[4]{Tammer Barkouki}
\author[5]{Brian J. Kirkwood}
\author[1,6]{Laura J. Brattain}
\affil[1]{Department of Electrical and Computer Engineering, University of Central Florida, Orlando, FL, USA}
\affil[2]{KnowledgeVis, LLC, Maitland, FL, USA}
\affil[3]{Ross School of Business, University of Michigan, Ann Arbor, MI, USA}
\affil[4]{Department of Mechanical and Aerospace Engineering, University of California, Davis, CA, USA}
\affil[5]{U.S. Army Institute of Surgical Research, JBSA Fort Sam Houston, Texas, USA}
\affil[6]{Department of Internal Medicine, University of Central Florida College of Medicine, Orlando, FL, USA}

\begin{document}
\maketitle
\thispagestyle{empty}
\pagestyle{empty}

\begin{abstract}

  Federated Learning (FL) provides a decentralized machine learning approach, where multiple devices or servers collaboratively train a model without sharing their raw data, thus enabling data privacy. This approach has gained significant interest in academia and industry due to its privacy-preserving properties, which are particularly valuable in the medical domain where data availability is often protected under strict regulations. A relatively unexplored area is the use of FL to fine-tune Foundation Models (FMs) for time series forecasting, potentially enhancing model efficacy by overcoming data limitation while maintaining privacy. In this paper, we fine-tuned time series FMs with Electrocardiogram (ECG) and Impedance Cardiography (ICG) data using different FL techniques. We then examined various scenarios and discussed the challenges FL faces under different data heterogeneity configurations. Our empirical results demonstrated that while FL can be effective for fine-tuning FMs on time series forecasting tasks, its benefits depend on the data distribution across clients. We highlighted the trade-offs in applying FL to FM fine-tuning.

\end{abstract}

\section{Introduction}
The field of Machine Learning (ML) has progressed significantly inhe past two decades. 
Recently, transformer-based architectures and the rise of Large Language Models (LLMs) have spurred increased interest in a new class of models known as Foundation Models (FMs). 
Trained in vast datasets, these models excel in a wide range of downstream tasks across various types of data and applications. 
Their ability to be fine-tuned for specific datasets further enhances their appeal.

Although much of the research on FMs has focused on Natural Language Processing (NLP) and Computer Vision (CV), there is growing interest in applying FMs to time-series forecasting \cite{chronos} \cite{lagllama} \cite{timesFM}. 
This application is particularly valuable in fields such as healthcare, where predicting future events can help physicians deliver preventive care, but often there is little data available for AI algorithm development. 
Therefore, FMs have the potential to have a significant impact in the healthcare industry through time series forecasting.

A major challenge in applying ML to biomedical data is the limited ability to share data due to privacy concerns. 
In healthcare, the sensitive nature of patient data makes this issue especially pressing. 
Federated learning (FL) has emerged as a promising decentralized solution, enabling ML models to be trained locally, thus alleviating privacy concerns \cite{fl}.

FL operates in a distributed manner among multiple clients with local data and a central server. The server distributes a "global" model to clients, which then train the model on their local data and send updates back to the server. The server aggregates these updates by averaging the results from clients (FedAvg) and refining the global model. The process often repeats itself until a performance criteria is met. This approach allows users to retain data privacy while benefiting from collaborative learning.

Under ideal conditions, FedAvg achieves good performance with minimal accuracy loss compared to centralized ML, while preserving data privacy. 
However, challenges arise when the network, resources or data distribution conditions are less than optimal \cite{ye2023heterogeneous}. 
Key issues include system heterogeneity, where clients have varying computational resources and network access, and statistical heterogeneity, where data across clients is not independent and identically distributed.

Although research has addressed potential solutions for both types of heterogeneity in FL, there is limited focus on applying FL to time series forecasting with FMs, particularly in the medical domain. 
Fine-tuning FMs presents unique challenges. FL for FM fine-tuning may offer only marginal improvements over local training on individual clients without shared updates, as FMs are pretrained on large, diverse datasets and may already exhibit strong generalization. 
In this paper, we examine some of the key challenges in fine-tuning FMs on ECG and ICG data across both Independent and Identically Distributed (IID), or homogeneous, and Non-Independent and Identically Distributed (non-IID), or heterogeneous, settings. 
Our evaluation leverages various FL techniques to address these challenges.
Our results show that while FL can outperform local fine-tuning in some cases, it may also underperform in others, highlighting the nuanced trade-offs involved.


\section{Related Works}
\textbf{Time Series Forecasting} is a well-established problem that involves using historical data, often referred to as ``context,'' to predict future values, known as the ``forecast.'' 
It has been extensively studied across various domains, including finance, healthcare, and weather prediction. 
Traditional approaches have relied on statistical methods, such as ARIMA, which use autoregression and moving averages to model time series data and are widely adopted for their simplicity and interpretability \cite{box2015time}.

In recent decades, neural network-based solutions have gained popularity, particularly with the advent of Convolutional Neural Networks (CNNs) \cite{cnn} and Long Short-Term Memory (LSTM) \cite{lstm} networks, which excel at capturing long-term dependencies in sequential data through their memory cells and gating mechanisms.
More recently, transformer-based architectures have been explored for time series forecasting, leveraging self-attention mechanisms to capture long-range dependencies efficiently, as demonstrated in prior works such as \cite{NEURIPS2019_6775a063}, which addresses locality and memory bottlenecks in transformers, and \cite{liu2022pyraformer}, which introduces pyramidal attention for low-complexity long-range modeling.
Foundation Models (FMs) are large-scale, pretrained machine learning models trained on large amounts of data at scale, which can be adapted (fine-tuned) to a wide range of downstream tasks across diverse domains \cite{bommasani2022opportunitiesrisksfoundationmodels}.
Initially popularized in NLP and CV, they are now being applied to time series data to overcome the limited application-specific data available. 
Notable works in this area include Amazon's Chronos \cite{chronos}, Google's timesFM \cite{timesFM}, and Lag-llama \cite{lagllama}, all of which have demonstrated the potential of FMs in time series forecasting.

\textbf{Federated Learning (FL)} has also seen substantial research interest in recent years, particularly in addressing the challenges of systems and statistical heterogeneity. 
Systems heterogeneity refers to the variation in computational and network resources among clients, while statistical heterogeneity deals with non-IID data across clients.
While systems heterogeneity is crucial, our work focuses primarily on statistical heterogeneity, which poses significant challenges in FL.


To address statistical heterogeneity, several methods have been proposed. 
FedProx introduces partial updates to handle resource constraints and reduce local model deviation from the global model \cite{fedprox}, with convergence analysis demonstrating its potential to mitigate statistical heterogeneity challenges.
FedOpt introduces federated versions of popular optimizers like Adam and Adagrad to mitigate the effects of non-IID data \cite{fedopt}. 
SCAFFOLD focuses on reducing variance between clients and the central server, thus improving model convergence in heterogeneous environments \cite{scaffold}.
Although these strategies have been developed to handle non-IID data within FL, they have largely focused on class imbalance and classification tasks rather than time series.

While a few works have investigated FL on time series data in the medical domain on problems such as arrhythmia classification using 12-lead ECG signals  \cite{10301542} and personal identification using vital signs data \cite{10.1007/978-3-031-49361-4_3}, they have not focused on time series forecasting. 
Several studies have applied FL to time series forecasting in domains such as base station traffic prediction \cite{PERIFANIS2023109950} and generalized benchmarks \cite{yuan2024tacklingdataheterogeneityfederated}, with a strong focus on non-IID data. 
However, they have not explored the use of FMs in the medical domain.
Our work is novel in that it explores the fine-tuning of pretrained FMs using FL for medical time series forecasting, with a particular focus on exploring FL under statistical heterogeneity.

\section{Methodology}
In this section, we will detail the methodology employed by our empirical analysis. 
Specifically, we will describe the datasets used, the different FL approaches that we evaluated, and the data partitioning approaches that we experimented.

\subsection{Federated Learning approaches}
\begin{figure}
  \includegraphics[scale=0.28]{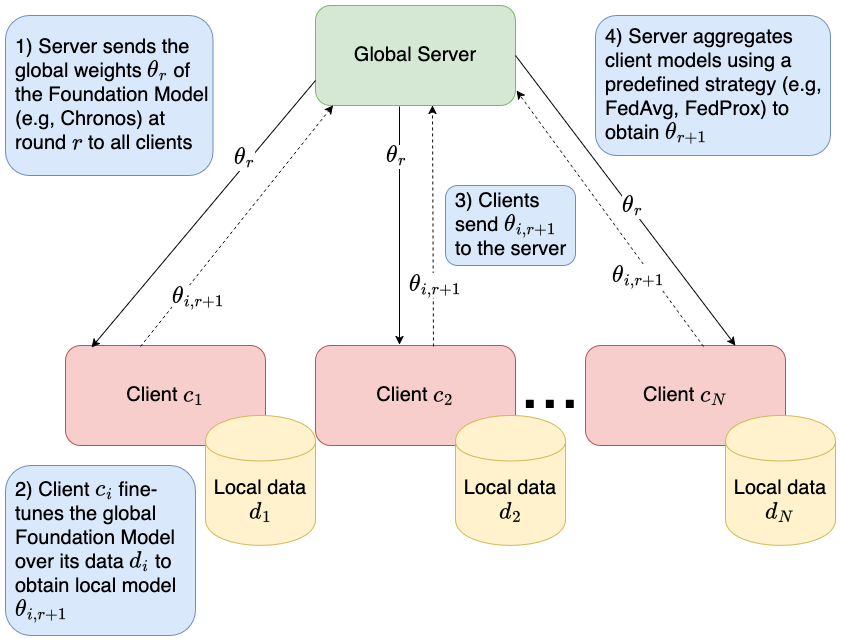}
  \caption{{Overview of fine-tuning a Foundation Model using Federated Learning at round \(r\). 
  The global server sends model weights \(\theta_r\) to clients, who update them locally on their data. 
  Clients return updated models \(\theta_{i,r+1}\), which the server aggregates using weighted averaging to obtain \(\theta_{r+1}\).
  }}
  \label{figs:fedavg_overview}
\end{figure}

\begin{figure}
  \includegraphics[scale=0.28]{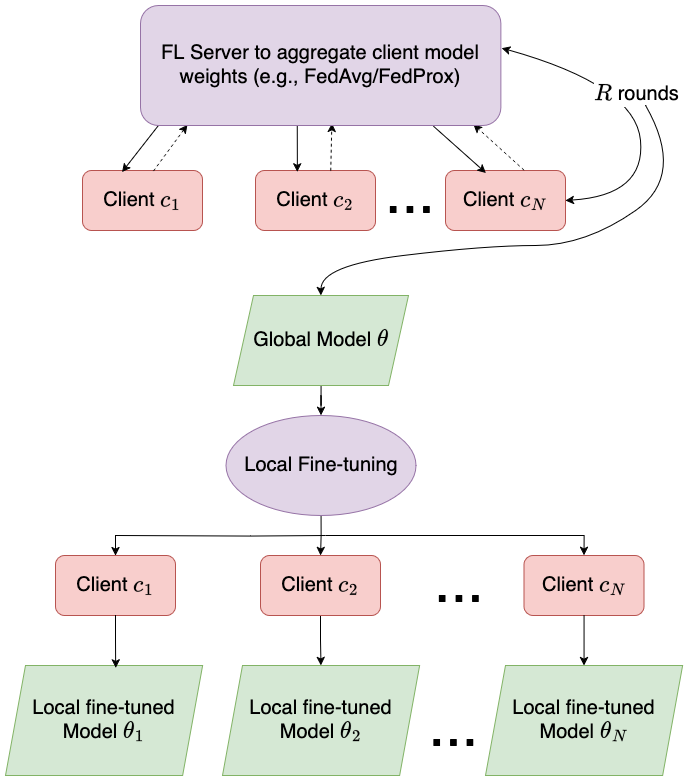}
  \caption{{An overview of a hybrid training architecture that combines Federated Learning and Local Adaptation.
  }}
  \label{figs:fed_la_overview}
\end{figure}

We evaluated three commonly used FL techniques: FedAvg \cite{fl}, FedProx \cite{fedprox}, and FL with local adaptation \cite{yu2022salvagingfederatedlearninglocal}.
Figure \ref{figs:fedavg_overview} illustrates the FL framework, while Figure \ref{figs:fed_la_overview} provides an overview of the local adaptation approach.
All three approaches fine-tuned the FM over many communication rounds with a goal to achieve generalizability (\(R=30\) in this paper).
We also evaluated the approach where the FM is fine-tuned on each local device separately, without creating a shared global model.
This is to compare against the results from the distributed FL frameworks and gain insight into trade-offs.

For the implementation of the FL algorithms, we used FlowerAI \cite{flower-ai}, an open source framework that streamlines the application of various FL algorithms in simulated settings.
The rest of this subsection provides a detailed background of the three FL approaches that we evaluated in this work.

\textbf{FedAvg} is the original FL model aggregation algorithm introduced in \cite{fl}.
The objective of the FedAvg algorithm is to find model parameters \(\theta\) such that the weighted average loss \( \frac{1}{|D|} \sum_{i=1}^{N} |d_i|f_i(\theta) \) is minimized. 
Here, \(f_i(\theta)\) is the loss over the local data \(d_i \in D\) of each client \( c_i \in C \). 
\(1 \leq i \leq N\) and \(N\) is the number of total clients sampled in each round \(r\), where \(1 \leq r \leq R\) and \(R\) is the number of total rounds in the FL process. 
\(|D|\) is the sum of the size of each client's local data, i.e \(|D|=\sum_{i=1}^{N}|d_i|\).

At the start of each round \(r + 1\), the central server sends the global model \(\theta_r\) to all or a subset of clients \(c_i \in C\). 
Each client \(c_i\) trains using this global model \(\theta_r\) and updates its local model at round \(r+1\) using a stochastic gradient step.
The local update for a single step is given by:
\[\theta_{i,r+1} =\theta_{i,r} - \eta \nabla_{\theta} f_i(\theta_{i,r})\]
\(\eta\) is the learning rate, and \(\nabla_{\theta}\) is the gradient of the loss with respect to the model weights. 
This represents a single local training step for client \(i\). 
In practice, clients perform multiple local updates before sending their models back to the server.

The global model, \(\theta_{r+1}\) is the weighted average of all the clients models, i.e \( \theta_{r+1} =  \frac{1}{|D|}  \sum_{i=1}^{N} |d_i|\theta_{i,r+1}\).
This could be written more simply as:
\[ \theta_{r+1} =\theta_r - \frac{\eta}{|D|} \sum_{i=1}^{N} |d_i| \nabla_{\theta_r}  f_i(\theta_{r}) \]

\textbf{FedProx} aims to mitigate the effects of statistical and system heterogeneity in FedAvg by introducing a proximal term in the local objective function, which constrains local updates to stay closer to the global model \cite{fedprox}. 
This, according to the authors, helps mitigate the impact of non-IID data and systems heterogeneity.
Even if the client's local model does not converge to its local data \(d_i \in D\), the authors state that by adding a proximal term, we can still leverage the local model updates.
More formally, the local loss function \(f'_i\) can be written with respect to the original loss function \(f_i\),
\[f'_i(\theta_{i,r+1}) = f_i(\theta_{i,r+1}) - \frac{\alpha}{2} | \theta_{i,r+1} -\theta_r |^2 \]
Here \(\alpha\) is a value between 0 and 1, and it controls the strength of the proximal term.
Recall that \(\theta_{i,r+1}\) is initialized with \(\theta_r\), and then gets updated using gradient steps. 
Hence, the second term is essentially a regularization term that helps the local model stay close to the global model.

\textbf{Federated Learning with Local Adaptation (Fed-LA)} is a slightly modified FL algorithm inspired by the idea that locally training models on clients can offer several benefits over training a single global model for all clients, as proposed in \cite{yu2022salvagingfederatedlearninglocal}.
The authors proposed an alternate hybrid strategy where a global model \(\theta_G\) is first obtained using \(R\) rounds of FL training.
This global model is then fine-tuned separately by clients \(c_i \in C\) on their local datasets \(d_i \in D\).
We made slight changes to this approach by adding a regularization term to the local objective function of each client during local fine-tuning (i.e after obtaining the global model using FL). 
This regularization term is with respect to the global model \(\theta_G\), to avoid the problem of catastrophic forgetting.
More formally, the updated client loss \(f''_i\) can be written as
\[ f''_i(\theta_i) = f_i(\theta_i) + |\theta_i - \theta_G|^2  \]
This is fairly similar to the local objective of FedProx, with the exception being that the global weights do not get updated after each round.
In this work, we obtained the global model by running FedAvg for 20 rounds before fine-tuning it separately on each client for another 10 rounds.
Hence, after completing 30 rounds, each client has its own local model, independent of the models from other clients.
For evaluation, each client tests on its own test dataset, and the results are combined using weighted averaging.

\subsection{Vital Sign Dataset}
One of the datasets that we utilized in this work is the vital signs dataset collected in \cite{Schellenberger2020-lb}. 
This dataset comprises of a total of 30 participants, with measurements taken using a non contact radar device and reference vital signs from a Task Force Monitor (TFM), which recorded ECG, impedance, impedance cardiogram (ICG), as well as non-invasive continuous blood pressure (BP). Two leads were recorded for ECG, one from the right arm (lead 1), the other from the left arm (lead 2).  
For this work, after assessing data quality, we decided to focus on ICG data (sampling rate of 500 Hz) and ECG Lead 2 (sampling rate of 500 Hz) due to its closer proximity to the heart. 
For each subject, measurements were taken under 1 or more scenarios. 
These scenarios included resting, Valsalva Manoeuvre (VM), apnea (holding breath), and the tilt table tests. 
In this work, we only consider the measurements taken under the resting scenario.

The dataset has 16 female and 14 male healthy participants, with an average BMI of 23.2 $ \pm $ 3.3 kg/
m\textsuperscript{2} and an average age of 30.7 $ \pm $ 9.9.

\subsection{PTB-XL Dataset}
To increase the applicability of our analysis, we also performed experiments on the Physikalisch-Technische Bundesanstalt Extra Large (PTB-XL) dataset collected in \cite{ptb-xl}.
This dataset consists of 21799 12-lead ECGs of 10-second each, sampled at 500 Hz.
There are 18869 patients, 52\% male and 48\% female, with ages ranging from 0 to 95 years (median 62 and interquantile range of 22). Data was collected from 52 different sites across hospitals and other healthcare organizations. 
This allowed us to simulate a real-world non-IID data distribution scenario, with each client having varying number of patient records and cardiac condition distributions,.
A previous study utilized the PTB-XL dataset for FL to classify arrhythmias using 12-lead ECG signals \cite{10.1007/978-3-031-49361-4_3}.
In contrast, our work focuses on time-series forecasting rather than classification.

\subsection{Data Partitions for Federated Learning}
As mentioned earlier, an important aspect to consider within FL is that of statistical heterogeneity, which refers to how the data is partitioned among different clients during the FL process. 
We thereby experimented under different data partitioning strategies, and performed an in-depth analysis of the performance of the different FL algorithms over these different data partitioning strategies. 
Three different data partitioning strategies were evaluated. 
A summary of the strategies is provided in Table \ref{tab:data_partitions}. 
For each strategy, we partitioned the data between 20 clients, and trained for 30 rounds.
Each client simulated a hospital-like entity.

\begin{table}
  \caption{\centering{A summary of the different data partitioning strategies used.}}
  \label{tab:data_partitions}
  \centering
  \begin{tabular}{l  l  l  l}
    \toprule 
    \textbf{Strategy}  & \textbf{No. of clients}   & \textbf{Data Distribution} & \textbf{Dataset} \\    
    \midrule
    Strategy \#1       & 20                         & IID                        & Vital Signs    \\
    \hline
    Strategy \#2       & 20                         & Non-IID                      & PTB-XL \\
    \hline
    Strategy \#3       & 20                         & Non-IID                      & Vital Signs \\
    
    \bottomrule
  \end{tabular}
\end{table}

\textbf{Strategy \#1:} For the first strategy, we maintained an IID data distribution among clients using the vital signs dataset.
We assigned 20 clients, each with ECG Lead 2 data from one or two patients, all recorded under the resting scenario. 
This setup ensured similar data distributions across clients, simulating the IID assumption. 
However, the small number of unique patients per client may limit model generalization, as some clients could learn patient-specific patterns rather than broader representations. 
While federated averaging mitigates this to some extent, further investigation is needed to assess the impact of patient distribution across clients. 
Despite this limitation, this setup provides a controlled environment to evaluate how well FL can fine-tune FMs on time-series medical data. 
All clients received an equal number of training samples to ensure a balanced comparison.

\textbf{Strategy \#2:} For the second strategy, we kept the data distribution among the clients non-IID.
We used the PTB-XL dataset for this experiment.
As aforementioned, the PTB-XL dataset has categorical labels indicating the site that each recording was taken at, information that could be used to simulate a real-world non-IID scenario.
For this experiment, we considered sites with more than 10 training recordings, and used 20 sites as clients for our experiments.
The clients had varying number of recordings. 
While some clients had more than 8500 training recordings, others had 10. 
We believe this added to the non-IID-ness of the client data.

\textbf{Strategy \#3:} To further explore the impact of data heterogeneity on fine-tuning FMs using FL, we performed another experiment targeting heterogeneity. 
Like the other two experiments, this experiment also had 20 clients. 
19 clients, just like Strategy \#1, had ECG Lead 2 data collected under the resting scenario. 
One client, however, trained exclusively on ICG data, collected under the resting scenario, with the task of predicting ICG signals instead of ECG. 
We included ICG to assess the FM’s ability to handle multiple modalities. 
However, our setup, where each client is restricted to a single modality, is a limitation, as a more realistic scenario would involve clients with access to multiple modalities.
Similar to Strategy \#1, all clients performed local fine-tuning for the same number of steps. 
Partitioning the data this way allowed us to further evaluate challenges associated in fine-tuning a time-series FM within the FL framework where there is a disparity in data distribution across clients. 
This partitioning strategy is different from Strategy \#2, where the distribution of each client is independent of every other client. 
Here, the distribution of 95\% of the clients (19 clients) is close to each other, while the distribution of 5\% (1 client) is different from the other clients.

We fine-tuned FMs on these partitioning strategies, and evaluated the resulting global models on test sets. 
We performed detailed comparisons between FedAvg, FedProx, Fed-LA, Local Training, and the zero-shot Chronos model (i.e., applying the Chronos model to our application-specific data as-is). 
Our results indicated a promising potential for fine-tuning FMs on time-series data using FL frameworks, as well as some key challenges pertaining to data heterogeneity that need attention and further investigation.

\subsection{Chronos}
Chronos, a class of FMs developed by Amazon, are time-series forecasting models pretrained on large corpus of time-series data. 
Chronos applies techniques similar to those used in LLMs, beginning by quantizing time-series context data and creating tokens through a tokenizer.
These tokens are fed into a large language model with an encoder-decoder or decoder-only architecture, which outputs predicted probabilities for the next time step.
These probabilities are used to generate the context token for the next time step, which is then converted into a numerical value, forming the model's prediction.
This process is repeated to predict any number of future time steps.
For our experiments, we used the default value in Chronos for context length, i.e., 512 past time steps, and forecast horizons up until 64 time steps.
Furthermore, while there are five different versions of Chronos, ranging from 8 million to 710 million parameters, we decided to utilize the model with approximately 8 million parameters (labeled tiny) for computation efficiency. 
All hyperparameter settings for the Chronos models were left as default.

\section{Evaluation}
In this section, we presented the empirical results from the experiments that we performed using the data partitioning scenarios described in the previous section.  

\subsection{Evaluation of Data Partition Strategy \#1}
For our first experiment, we simulated the scenario where the resting data of all 30 patients is divided among 20 clients. 
During each round, each client trained the model on its local data for 400 steps and 64 batch size.

To evaluate different FL approaches, we train using each approach for 30 rounds. 
After each round, clients test the global model on their local data separate from their training sets. 
The results are then averaged by the global server using weighted averaging, where weights are the size of the clients' test sets.
For Fed-LA, FedAvg's global weights after round 20 were used to fine-tune the model locally for 10 rounds.

For the local training approach, we kept all hyperparameters the same as the FL experiments, except that the model weights are not shared with a central sever. 
We also evaluated the zero-shot Chronos model on the test set.

\begin{figure}
  \includegraphics[scale=0.45]{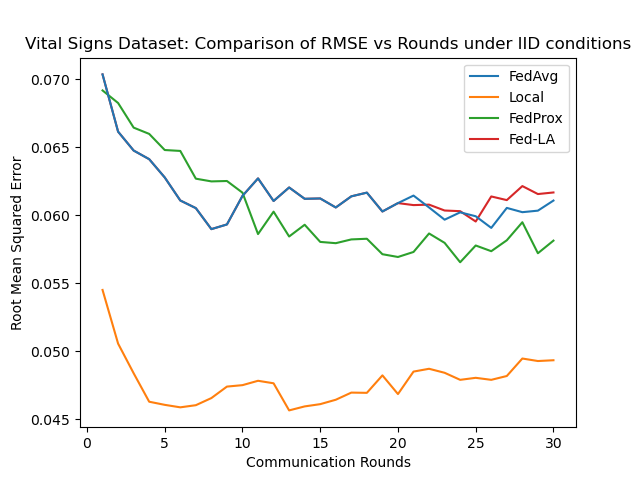}
  \includegraphics[scale=0.45]{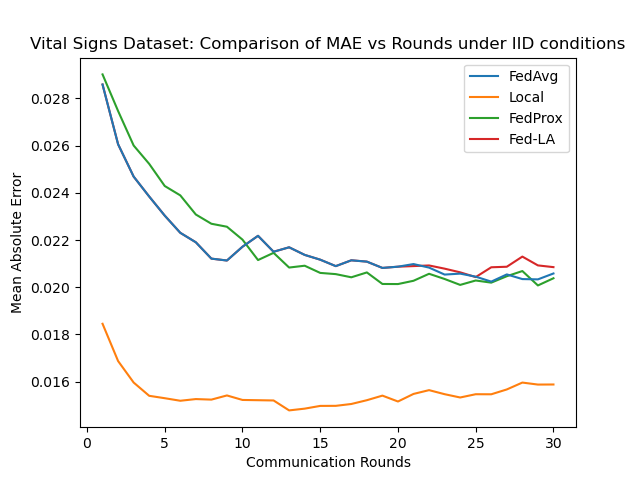}
  \caption{{Comparisons of FedAvg, FedProx, Fed-LA, and local fine-tuning models in terms of RMSE and MAE against the number of rounds when the data distribution amongst clients is IID. 
  All FL algorithms were outperformed by the locally fine-tuned models.}}
  \label{figs:homo_comparison}
\end{figure}

Table \ref{tab:homo_comparison} shows a comparison of the performance of all the approaches. 
Multiple metrics were used to evaluate the performance - the Root Mean Squared Error (RMSE), Mean Absolute Error (MAE), and symmetric Mean Percentage Error (sMAPE). 
These errors refer to the error observed when evaluating the 64-step forecast. 
We also performed another comparison where we compared the performance of the different approaches against the number of rounds. 
Figure \ref{figs:homo_comparison} shows this comparison in terms of RMSE and MAE. 

\begin{table}
  \caption{{Strategy \#1: Comparison of three FL approaches against zero-shot and local fine-tuning for time-series forecasting with 64-point forecast horizons.}}
  \label{tab:homo_comparison}
  \centering
  \begin{tabular}{l l l l}
    \toprule 
    Approach                   & RMSE      & MAE    & sMAPE (\%)  \\
    \midrule
    FedAvg                  & 0.061    & 0.021   & 6.73       \\
    FedProx                 & 0.058    & 0.020   & 6.81       \\
    Fed-LA                   & 0.062    & 0.021   & 6.83       \\
    Zero-shot               & 0.091    & 0.039   & 12.67      \\
    Local Fine-tuning       & 0.049    & 0.016   & 5.43       \\
    \bottomrule
  \end{tabular}
\end{table}

Based on Figure \ref{figs:homo_comparison} and Table \ref{tab:homo_comparison}, we observe that FL effectively fine-tunes the pre-trained Chronos FM for ECG time series data in this case.
FL outperforms zero-shot across all metrics and forecast horizons, showing that FL can be a useful approach. 
However, the local fine-tuning approach outperforms all the FL approaches, performing better across all metrics. 
While the performance gap is not very significant, it does mean that clients in this particular scenario, overall, do not benefit from FL.

\subsection{Evaluation of Data Partition Strategy \#2}
For this experiment, we used the PTB-XL dataset to distribute the data among clients.
Each client represented a site where the data was originally collected.
There were 20 clients, and the number of samples between the clients varied from 11 to 8910 training recordings. 
The number of training steps for each client was kept the same as the client's training recordings, to simulate the scenario where there is a data imbalance among clients.
Testing was performed using the same approach as Strategy \#1, where each client (each site, in this case) has a 70/30 train/test split.

\begin{figure}
  \includegraphics[scale=0.45]{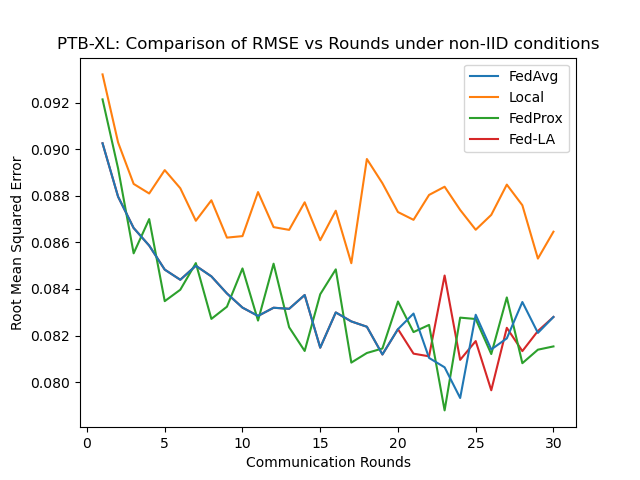}
  \includegraphics[scale=0.45]{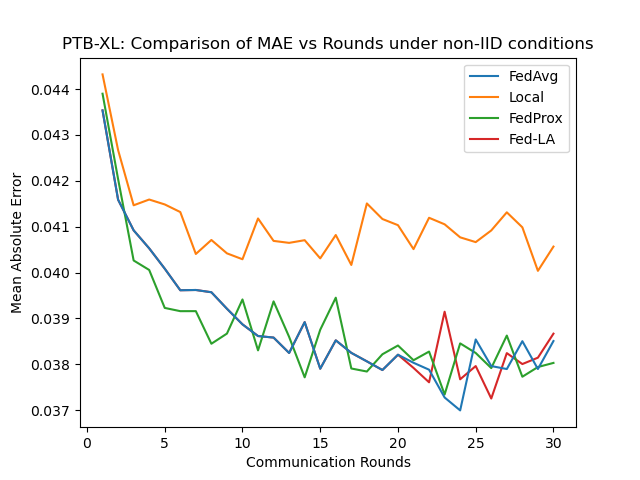}
  \caption{{Comparisons of FedAvg, FedProx, Fed-LA, and local fine-tuning models in terms of RMSE and MAE against the number of rounds when the data distribution amongst clients is non-IID. 
  FL algorithms outperformed zero-shot and local fine-tuning.}}
  \label{figs:ptbxl_niid_comparison}
\end{figure}

Table \ref{tab:strategy_2_summary} shows a comparison of the average performance on 64-point forecasts after all rounds of fine-tuning. 
We include the same metrics as the previous experiment. 
Figure \ref{figs:ptbxl_niid_comparison} visualizes the performance of the model with different fine-tuning approaches in terms of RMSE and MAE in relation to the communication rounds. 

\begin{table}
  \caption{{Strategy \#2: Comparison of three FL approaches against zero-shot and local fine-tuning for time-series forecasting with 64-point forecast horizons.}}
  \label{tab:strategy_2_summary}
  \centering
  \begin{tabular}{l l l l}
    \toprule 
    Approach            & RMSE      & MAE     & sMAPE (\%)  \\
    \midrule
    
    FedAvg             & 0.083     & 0.039   & 14.0       \\
    FedProx            & 0.082     & 0.038   & 13.9      \\
    Local              & 0.086     & 0.041   & 14.9       \\
    Fed-LA              & 0.083     & 0.038   & 14.0       \\
    Zero-shot          & 0.146     & 0.076    & 23.6        \\
    \bottomrule
  \end{tabular}
\end{table}

Table \ref{tab:strategy_2_summary} and Figure \ref{figs:ptbxl_niid_comparison} show FL algorithms performing better than the local fine-tuning and zero-shot approaches over the PTB-XL dataset with non-IID partitioning.
This is different than the evaluation over Strategy \#1, where the local fine-tuning approach provided the best performance.
However, we note that the performance difference between FL and local fine-tuning may not be significant.
Moreover, Fed-LA's performance aligns with the first experiment, as local fine-tuning did not significantly alter the model's convergence.

\subsection{Evaluation of Data Partition Strategy \#3}
While the previous experiment tested a partitioning strategy where the data distribution among all the clients was fairly different (since all collected their data independently of each other), we performed another experiment with a non-IID partitioning strategy.
We used the vital signs dataset, and similar to Strategy \#1, we divided data between 20 clients, each locally training for the same number of steps per round (400 steps).  
In this partitioning strategy, to test a scenario where there is an extreme data imbalance (e.g., 5\% of the data is ICG data), one of the clients has ICG data instead of ECG Lead 2 data. 
All clients, except for one, follow an IID distribution.
FMs are generally trained on large amounts of data, and should be able to perform well on different tasks.
Hence, in the ideal scenario, we want to use the same model for ICG and ECG.

Table \ref{tab:strategy_3_summary} shows a comparison of the performance over the 64 time-step forecast when evaluating over both ECG and ICG data using different FL techniques. It can be observed that the performance of all the fine-tuned FL models was worse than the locally fine-tuned model when forecasting both ECG and ICG signals. 
This performance gap is especially significant when forecasting ICG signal, with local fine-tuning significantly outperforming the FL fine-tuning approaches.
This indicates a significant challenge in fine-tuning FMs for time series forecasting when there is underrepresented data across clients. 
For ECG data, we see that FL resulted much better performance, outperforming the zero-shot model significantly.
In contrast, for ICG data, the FL model's performance was fairly similar to the zero-shot model.
This discrepancy in results across the two types of data highlights the impact of data characteristics and data distribution on the performance of FMs fine-tuned using FL. 


\begin{table}[htbp]
\caption{Strategy \#3: Comparison of the performance of different FL approaches against zero-shot and local fine-tuning. The table breaks down the performance over ECG and ICG data for 64-point forecast horizons.}
\centering
\resizebox{\columnwidth}{!}{\begin{tabular}{@{}lSSSSSSSS@{}}
    \toprule
    \textbf{Approach} &
      \multicolumn{3}{c}{\textbf{ECG}} &
      \multicolumn{3}{c}{\textbf{ICG}}  \\
      & {RMSE} & {MAE} & {sMAPE} & {RMSE} & {MAE} & {sMAPE} \\
      \cmidrule(r){2-4}\cmidrule(l){5-7}
    
    FedAvg    & 0.056 &  0.020 & 6.78\%  & 0.065 & 0.039 & 10.63\%  \\
    FedProx   & 0.058 &  0.021 & 7.14\%  & 0.074 & 0.045 & 12.27\%  \\
    Local     & 0.045 &  0.016 & 5.30\%  & 0.041 & 0.025 & 6.94\%  \\
    Fed-LA     & 0.057 &  0.021 & 6.81\%  & 0.068 & 0.040 & 11.08\%  \\
    Zero-shot & 0.091 & 0.039  & 12.67\% & 0.061 & 0.039 & 10.46\%  \\
    \bottomrule
  \end{tabular}}
  \label{tab:strategy_3_summary}
\end{table}

Overall, our evaluation results across the three partitioning strategies highlight the nuanced differences between using federated learning (FL) for fine-tuning foundation models (FMs) versus local fine-tuning. 
We find that local fine-tuning outperforms traditional FL algorithms when the data distribution of individual clients closely aligns with the overall distribution across all clients.
However, in scenarios where each client's local data distribution is independent of others (Strategy \#2), FL techniques outperformed local fine-tuning, albeit by a small margin.
Additionally, our results with Fed-LA, a hybrid approach combining federated learning with local adaptation, reveal that the fine-tuning process in FL can impact the model's convergence, making it more difficult for the model to effectively adapt to clients' local data during the subsequent fine-tuning stage.

\section{Conclusion}
In this paper, we evaluated the use of FL to fine-tuned FMs for medical time series forecasting, focusing on ECG and ICG data. Leveraging FMs within an FL framework has the potential to reduce the need for large, centralized medical datasets while preserving privacy. 
However, fine-tuning FMs in FL settings introduces unique challenges, particularly due to statistical heterogeneity across clients.

Through a series of experiments with different data partitioning strategies, we demonstrated that FL can effectively fine-tune FMs, leading to significant improvements over zero-shot models.
However, our findings also highlighted that the effectiveness of FL is highly dependent on the underlying data distribution. 
In scenarios where most clients have IID data, local fine-tuning outperformed FL, which could discourage clients from participating in FL due to minimal added benefit. This challenge is particularly relevant to FMs, as they are already well-generalized from pretraining on large, diverse datasets.

Despite these limitations, FL remains a promising approach for adapting FMs in privacy-sensitive medical applications. 
Our work underscores the need for advanced techniques to address statistical heterogeneity in FL, ensuring more robust and generalizable models for healthcare.

Future research should explore methods that strike a balance between localized fine-tuning and collaborative learning across clients. 
While we experimented with a hybrid approach where a globally fine-tuned FM was later adapted locally, further convergence analysis is needed to better understand its impact. 
Additionally, meta-learning approaches, such as Model-Agnostic Meta-Learning (MAML) \cite{10.5555/3305381.3305498}, offer a compelling direction. 
Personalized Federated Learning (PFL) techniques \cite{NEURIPS2020_24389bfe}, which integrate meta-learning with FL, could enhance adaptability by enabling FMs to be fine-tuned on local clients with better convergence guarantees. Such approaches may offer a more effective trade-off between FL and local fine-tuning, improving both performance and participation incentives.

By systematically evaluating these strategies, we can further enhance the robustness of FL across diverse medical data distributions. 
Overall, integrating FMs into FL frameworks has the potential to advance medical AI by reducing data requirements while preserving privacy, ultimately contributing to improved patient outcomes.

\bibliographystyle{plain}
\bibliography{references}

\begin{thebibliography}{10}

\bibitem{chronos}
Abdul~Fatir Ansari, Lorenzo Stella, Caner Turkmen, Xiyuan Zhang, Pedro Mercado, Huibin Shen, Oleksandr Shchur, Syama~Sundar Rangapuram, Sebastian~Pineda Arango, Shubham Kapoor, et~al.
\newblock Chronos: Learning the language of time series.
\newblock {\em arXiv preprint arXiv:2403.07815}, 2024.

\bibitem{flower-ai}
Daniel~J. Beutel, Taner Topal, Akhil Mathur, Xinchi Qiu, Javier Fernandez-Marques, Yan Gao, Lorenzo Sani, Kwing~Hei Li, Titouan Parcollet, Pedro Porto~Buarque de~Gusmão, and Nicholas~D. Lane.
\newblock Flower: A friendly federated learning research framework, 2022.

\bibitem{bommasani2022opportunitiesrisksfoundationmodels}
Rishi Bommasani, Drew~A. Hudson, Ehsan Adeli, Russ Altman, Simran Arora, Sydney von Arx, Michael~S. Bernstein, Jeannette Bohg, Antoine Bosselut, Emma Brunskill, et~al.
\newblock On the opportunities and risks of foundation models, 2022.

\bibitem{box2015time}
George~EP Box, Gwilym~M Jenkins, Gregory~C Reinsel, and Greta~M Ljung.
\newblock {\em Time series analysis: forecasting and control}.
\newblock John Wiley \& Sons, 2015.

\bibitem{timesFM}
Abhimanyu Das, Weihao Kong, Rajat Sen, and Yichen Zhou.
\newblock A decoder-only foundation model for time-series forecasting, 2024.

\bibitem{NEURIPS2020_24389bfe}
Alireza Fallah, Aryan Mokhtari, and Asuman Ozdaglar.
\newblock Personalized federated learning with theoretical guarantees: A model-agnostic meta-learning approach.
\newblock In H.~Larochelle, M.~Ranzato, R.~Hadsell, M.F. Balcan, and H.~Lin, editors, {\em Advances in Neural Information Processing Systems}, volume~33, pages 3557--3568. Curran Associates, Inc., 2020.

\bibitem{10.5555/3305381.3305498}
Chelsea Finn, Pieter Abbeel, and Sergey Levine.
\newblock Model-agnostic meta-learning for fast adaptation of deep networks.
\newblock In {\em Proceedings of the 34th International Conference on Machine Learning - Volume 70}, ICML'17, page 1126–1135. JMLR.org, 2017.

\bibitem{lstm}
Sepp Hochreiter and J\"{u}rgen Schmidhuber.
\newblock Long short-term memory.
\newblock {\em Neural Comput.}, 9(8):1735–1780, nov 1997.

\bibitem{10301542}
Tae-Ho Hwang, Jingyao Shi, and Kangyoon Lee.
\newblock Enhancing privacy-preserving personal identification through federated learning with multimodal vital signs data.
\newblock {\em IEEE Access}, 11:121556--121566, 2023.

\bibitem{10.1007/978-3-031-49361-4_3}
Daniel~Mauricio Jimenez~Gutierrez, Hafiz~Muuhammad Hassan, Lorella Landi, Andrea Vitaletti, and Ioannis Chatzigiannakis.
\newblock Application of federated learning techniques for arrhythmia classification using 12-lead ecg signals.
\newblock In Ioannis Chatzigiannakis and Ioannis Karydis, editors, {\em Algorithmic Aspects of Cloud Computing}, pages 38--65, Cham, 2024. Springer Nature Switzerland.

\bibitem{scaffold}
Sai~Praneeth Karimireddy, Satyen Kale, Mehryar Mohri, Sashank Reddi, Sebastian Stich, and Ananda~Theertha Suresh.
\newblock Scaffold: Stochastic controlled averaging for federated learning.
\newblock In {\em International Conference on Machine Learning}, pages 5132--5143. PMLR, 2020.

\bibitem{cnn}
Yann LeCun and Yoshua Bengio.
\newblock {\em Convolutional networks for images, speech, and time series}, page 255–258.
\newblock MIT Press, Cambridge, MA, USA, 1998.

\bibitem{NEURIPS2019_6775a063}
Shiyang Li, Xiaoyong Jin, Yao Xuan, Xiyou Zhou, Wenhu Chen, Yu-Xiang Wang, and Xifeng Yan.
\newblock Enhancing the locality and breaking the memory bottleneck of transformer on time series forecasting.
\newblock In H.~Wallach, H.~Larochelle, A.~Beygelzimer, F.~d\textquotesingle Alch\'{e}-Buc, E.~Fox, and R.~Garnett, editors, {\em Advances in Neural Information Processing Systems}, volume~32. Curran Associates, Inc., 2019.

\bibitem{fedprox}
Tian Li, Anit~Kumar Sahu, Manzil Zaheer, Maziar Sanjabi, Ameet Talwalkar, and Virginia Smith.
\newblock Federated optimization in heterogeneous networks.
\newblock {\em Proceedings of Machine learning and systems}, 2:429--450, 2020.

\bibitem{liu2022pyraformer}
Shizhan Liu, Hang Yu, Cong Liao, Jianguo Li, Weiyao Lin, Alex~X. Liu, and Schahram Dustdar.
\newblock Pyraformer: Low-complexity pyramidal attention for long-range time series modeling and forecasting.
\newblock In {\em International Conference on Learning Representations}, 2022.

\bibitem{fl}
Brendan McMahan, Eider Moore, Daniel Ramage, Seth Hampson, and Blaise~Aguera y~Arcas.
\newblock Communication-efficient learning of deep networks from decentralized data.
\newblock In {\em Artificial intelligence and statistics}, pages 1273--1282. PMLR, 2017.

\bibitem{PERIFANIS2023109950}
Vasileios Perifanis, Nikolaos Pavlidis, Remous-Aris Koutsiamanis, and Pavlos~S. Efraimidis.
\newblock Federated learning for 5g base station traffic forecasting.
\newblock {\em Computer Networks}, 235:109950, 2023.

\bibitem{lagllama}
Kashif Rasul, Arjun Ashok, Andrew~Robert Williams, Hena Ghonia, Rishika Bhagwatkar, Arian Khorasani, Mohammad Javad~Darvishi Bayazi, George Adamopoulos, Roland Riachi, Nadhir Hassen, Marin Biloš, Sahil Garg, Anderson Schneider, Nicolas Chapados, Alexandre Drouin, Valentina Zantedeschi, Yuriy Nevmyvaka, and Irina Rish.
\newblock Lag-llama: Towards foundation models for probabilistic time series forecasting, 2024.

\bibitem{fedopt}
Sashank Reddi, Zachary Charles, Manzil Zaheer, Zachary Garrett, Keith Rush, Jakub Kone{\v{c}}n{\`y}, Sanjiv Kumar, and H~Brendan McMahan.
\newblock Adaptive federated optimization.
\newblock {\em arXiv preprint arXiv:2003.00295}, 2020.

\bibitem{Schellenberger2020-lb}
Sven Schellenberger, Kilin Shi, Tobias Steigleder, Anke Malessa, Fabian Michler, Laura Hameyer, Nina Neumann, Fabian Lurz, Robert Weigel, Christoph Ostgathe, and Alexander Koelpin.
\newblock A dataset of clinically recorded radar vital signs with synchronised reference sensor signals.
\newblock {\em Scientific Data}, 7(1):291, September 2020.

\bibitem{ptb-xl}
Patrick Wagner, Nils Strodthoff, Ralf-Dieter Bousseljot, Dieter Kreiseler, Fatima~I. Lunze, Wojciech Samek, and Tobias Schaeffter.
\newblock Ptb-xl, a large publicly available electrocardiography dataset.
\newblock {\em Scientific Data}, 7(1):154, May 2020.

\bibitem{ye2023heterogeneous}
Mang Ye, Xiuwen Fang, Bo~Du, Pong~C Yuen, and Dacheng Tao.
\newblock Heterogeneous federated learning: State-of-the-art and research challenges.
\newblock {\em ACM Computing Surveys}, 56(3):1--44, 2023.

\bibitem{yu2022salvagingfederatedlearninglocal}
Tao Yu, Eugene Bagdasaryan, and Vitaly Shmatikov.
\newblock Salvaging federated learning by local adaptation, 2022.

\bibitem{yuan2024tacklingdataheterogeneityfederated}
Wei Yuan, Guanhua Ye, Xiangyu Zhao, Quoc Viet~Hung Nguyen, Yang Cao, and Hongzhi Yin.
\newblock Tackling data heterogeneity in federated time series forecasting, 2024.

\end{thebibliography}

\end{document}